\documentclass{article}
\usepackage{spconf,amsmath,graphicx}
\usepackage{amssymb}
\usepackage{bm}
\usepackage{url}
\usepackage{booktabs}

\usepackage{color}
\title{A GRAPH-CNN FOR 3D POINT CLOUD CLASSIFICATION}
\name{Yingxue Zhang and Michael Rabbat}
\address{McGill University\\
    Montreal, Canada}

\begin{document}
\ninept
\maketitle
\begin{abstract}
\textit{Graph convolutional neural networks} (Graph-CNNs) extend traditional CNNs to handle data that is supported on a graph. Major challenges when working with data on graphs are that the support set (the vertices of the graph) do not typically have a natural ordering, and in general, the topology of the graph is not regular (i.e., vertices do not all have the same number of neighbors). Thus, Graph-CNNs have huge potential to deal with 3D point cloud data which has been obtained from sampling a manifold. In this paper we develop a Graph-CNN for classifying 3D point cloud data, called \textit{PointGCN}\footnote{Code is available at \url{https://github.com/maggie0106/Graph-CNN-in-3D-Point-Cloud-Classification}}. The architecture combines localized graph convolutions with two types of graph downsampling operations (also known as pooling). By the effective exploration of the point cloud local structure using the Graph-CNN, the proposed architecture achieves competitive performance on the 3D object classification benchmark ModelNet, and our architecture is more stable than competing schemes.

\end{abstract}


\section{Introduction}
\label{sec:intro}
With the advent of very large datasets and improved computational capabilities, methods using \emph{convolutional neural networks} (CNNs) now achieve state-of-the-art performance on a variety of tasks, including speech recognition and image classification. Many emerging applications give rise to data that may be viewed as being supported on the vertices of a graph, and field of \emph{graph signal processing} (GSP) has developed filtering and other operations on graph signals~\cite{DBLP:journals/spm/ShumanNFOV13,Sandryhaila2014}. Data may either be naturally sampled on the vertices or edges of a graph (e.g., flows on a transportation network), or the data may simply be unstructured and a graph is imposed to capture the manifold structure underlying the data (e.g., the 3D point clouds considered in this paper). Unlike the domains encountered in more traditional signal processing (e.g., 1D time-series, 2D images), general graph topologies do not have the same regularity or symmetries, and so there is not a unique, well-defined notion of convolution on a graph. This has motivated researchers to develop a variety of approaches to convolutions on graphs, which can then be applied in graph-CNNs and other graph-based signal processing architectures.

Bruna et al.~\cite{DBLP:journals/corr/BrunaZSL13,DBLP:journals/corr/HenaffBL15} first proposed the idea of using a graph convolution defined in the graph spectral domain together with a graph multiresolution clustering approach to achieve pooling/downsampling. Defferrard et al.~\cite{DBLP:conf/nips/DefferrardBV16} propose a fast localized convolution operation by leveraging the recursive form of Chebyshev polynomials to both avoid explicitly calculating the Fourier graph basis and to allow the number of learnable filter coefficients to be independent of the graph size. Atwood and Towsley~\cite{DBLP:conf/nips/AtwoodT16} use a similar localized filtering idea but define the convolution process directly in the spatial domain by searching the receptive filed at different scales using random walk. 

Graph kernels have also been applied to the graph classification task \cite{DBLP:conf/icml/KondorL02, DBLP:journals/jmlr/ShervashidzeSLMB11} which aims to classify a graph based on its topology, as opposed to classifying or otherwise processing signals on a graph. However, Graph kernels suffer from quadratic training complexity in the number of graphs \cite{DBLP:journals/jmlr/ShervashidzeSLMB11}.

The formulation of Graph-CNNs opens up a range of applications. Defferrard et al.~\cite{DBLP:conf/nips/DefferrardBV16} validate their model on an image classification task and demonstrate the effectiveness of Graph-CNNs. Kipf and Welling \cite{DBLP:journals/corr/KipfW16} study the application of the Graph-CNNs to semi-supervised learning. In this paper, we explore the application of the Graph-CNNs in 3D point cloud data.

GSP techniques have been applied to process 3D point cloud data, such as that obtained by \textit{light detection and ranging} (LiDAR) sensors. Rather than binning point clouds into voxels, graph-based approaches fit a graph with one vertex for each point and edges between nearby points, and then operate on the graph. The effectiveness of GSP for processing 3D point cloud data has been demonstrated in applications such as data visualization, in-painting, and compression~\cite{DBLP:conf/icassp/ChenTFVK17, DBLP:journals/corr/ChenTFVK17,DBLP:journals/spm/LozesEL15,DBLP:journals/tip/ThanouCF16}.

In this work, we propose a Graph-CNN architecture called PointGCN for classifying 3D point cloud data by exploring its local structure encoded in the constructed graph. Unlike most previous Graph-CNNs, in this setting both the signals and the graph structure vary from input to input. The proposed architecture uses existing graph convolution operation together with two types of specifically designed pooling layers for point cloud data. The architecture learns a latent signature summarizing each point cloud at different receptive fields.

We achieve an average classification accuracy comparable to the state-of-the-art on the ModelNet benchmark, and the variance of the proposed approach is substantially lower than existing point-based classification methods.

\section{Problem Statement}
\label{sec:format}
We consider a classification problem where we are given $m$ labeled training instances $\{(X_j, y_j)\}$, each composed of an input $X_j \in \mathcal{X}$ and an output $y_j \in \mathcal{Y}$. Our goal is produce a function $y = f(X)$ to predict the output $y$ associated with a new, unseen input $X$. For point-based 3D classification problem, we consider the case where the output space $\mathcal{Y}$ is finite (the classes), and each input $X_j$ is a set of $n$ points, $\{x_{j,1}, \dots, x_{j,n}\} \subset \mathbb{R}^3$.

Previous work has taken different approaches to classifying 3D point clouds~\cite{DBLP:conf/cvpr/WuSKYZTX15,DBLP:conf/iros/MaturanaS15,DBLP:journals/corr/GrahamM17,DBLP:conf/iccv/SuMKL15,DBLP:journals/corr/QiSMG16,DBLP:conf/cvpr/SimonovskyK17, qi2017pointnet++, DBLP:conf/nips/deepsets}, including rendering and processing a collection of 2D images (projections of the points onto an image plane from different perspectives), or binning the points into voxels. However, the former involves extensive data augmentation, preprocessing and heavy computation, and the latter introduces discretization error as well as use a relatively sparse format to represent the 3D data. Directly using the point cloud format to tackle the 3D classification problem is an area of research interest. PointNet~\cite{DBLP:journals/corr/QiSMG16} is the pioneer in this field, which uses spatial transformation network and symmetric functions to learn the global representation of each point cloud object. Deep Sets \cite{DBLP:conf/nips/deepsets} constructs similar universal approximators to achieve invariance to permutation within each point cloud. However, the lack of local structure exploration leads to the loss of local context of each point. The following work, PointNet++ \cite{qi2017pointnet++}, tries to address this problem, but each point is still treated individually in the constructed local clusters. We instead take a graph-based approach, fitting a graph and learning a mapping from the space of graphs and its corresponding graphs signal to classes. By doing so, the local geometric details are encoded in the pairwise distance between center point and its neighbors.

\section{Methodology}

Given a set of points $X = \{x_i\}_{i=1}^n \subset \mathbb{R}^3$, we first fit a (symmetrized) $k$-nearest neighbor graph to the points and then weight each edge using a Gaussian kernel: for $i,j = 1,\dots,n,$

\begin{equation}
W_{i,j}=
\begin{cases}
\exp(- \| x_i - x_j \|^2 / \sigma^2) & \text{ if } j \in \mathcal{A}_k(i) \\
0 & \text{ otherwise,}
\end{cases}
\label{graph_setup}
\end{equation}
where $\mathcal{A}_k(i)$ is the set of $k$ nearest neighbors of vertex $i$. 

We will also use the coordinates as graph signals. Let $x_i = [x_i^{(1)}, x_i^{(2)}, x_i^{(3)}]^T$, and let $\bm{x}^{(c)} \in \mathbb{R}^n$ denote the vector with $i$th entry equal to $x_i^{(c)}$. Then $\bm{x}^{(c)}$ can be seen as a signal on the graph (one value per vertex). The associated coordinate vectors $\bm{x}^{(1)}, \bm{x}^{(2)}, \bm{x}^{(3)} \in \mathbb{R}^{n\times3}$ will serve as graph signals.

\subsection{Graph Signal Processing}

We briefly review notions of graph filtering and convolution before describing the proposed architecture. Given the symmetric weighted adjacency matrix $W \in \mathbb{R}^{n \times n}$ of a graph, let $L = I_n - D^{-1/2} W D^{-1/2}$ denote the normalized Laplacian matrix. 
 A linear vertex-domain graph filter with coefficients $\alpha_0, \dots, \alpha_K$ transforms one graph signal, $\bm{x}$, to another, $\bm{y}$, via 
$\bm{y} = h_\alpha(L) \bm{x} = \sum_{k=0}^K \alpha_k L^k \bm{x}.$
It can also be convenient to represent or approximate filters in terms of Chebyshev polynomials of $L$ \cite{hammond2011wavelets}, 
\begin{equation}
\bm{y} = g_\theta(L) \bm{x} = \sum_{k=0}^{K} \theta_k T_k(L) \bm{x}, \label{cheby}
\end{equation}
defined recursively via $T_0(L) = I$, $T_1(L) = L$, and for $k \ge 2$,
\[
T_k(L) = 2 L T_{k-1}(L) - T_{k-2}(L).
\]
Graph-CNNs involve multiple such filters where the coefficients $\{\alpha_k\}$ or ${\theta_k}$ are learned from data.

Graph filters have a spectral interpretation, in terms of the eigendecomposition $L = U \Lambda U^{T}$ of the Laplacian. We have
\[
\bm{y} = h_\alpha(L) \bm{x} = U h_\alpha(\Lambda) U^T \bm{x} = U h_\alpha(\Lambda) \widehat{\bm{x}},
\]
so the eigenvectors $U^T$ serve as a graph spectral basis, $\widehat{\bm{x}}$ is the vector of graph spectral coefficients of $\bm{x}$, and the filter $h_\alpha(\cdot)$ acts entry-wise on the eigenvalues $\Lambda$ (with an identical interpretation possible in terms of $g_\theta$). The eigenvectors of the Laplacian are known to correspond to low- or high-variation over the graph proportional to the corresponding eigenvalue~\cite{DBLP:journals/spm/ShumanNFOV13}.

Since different types of filters emphasize different structural characteristics, in this work we take the approach of learning graph filter bank in order to obtain features which can optimize the downstream point cloud classification task.

\subsection{Proposed Graph-CNN Architecture for Point Cloud Classification}

\begin{figure*}[t]
\centering
\includegraphics[width=13cm]{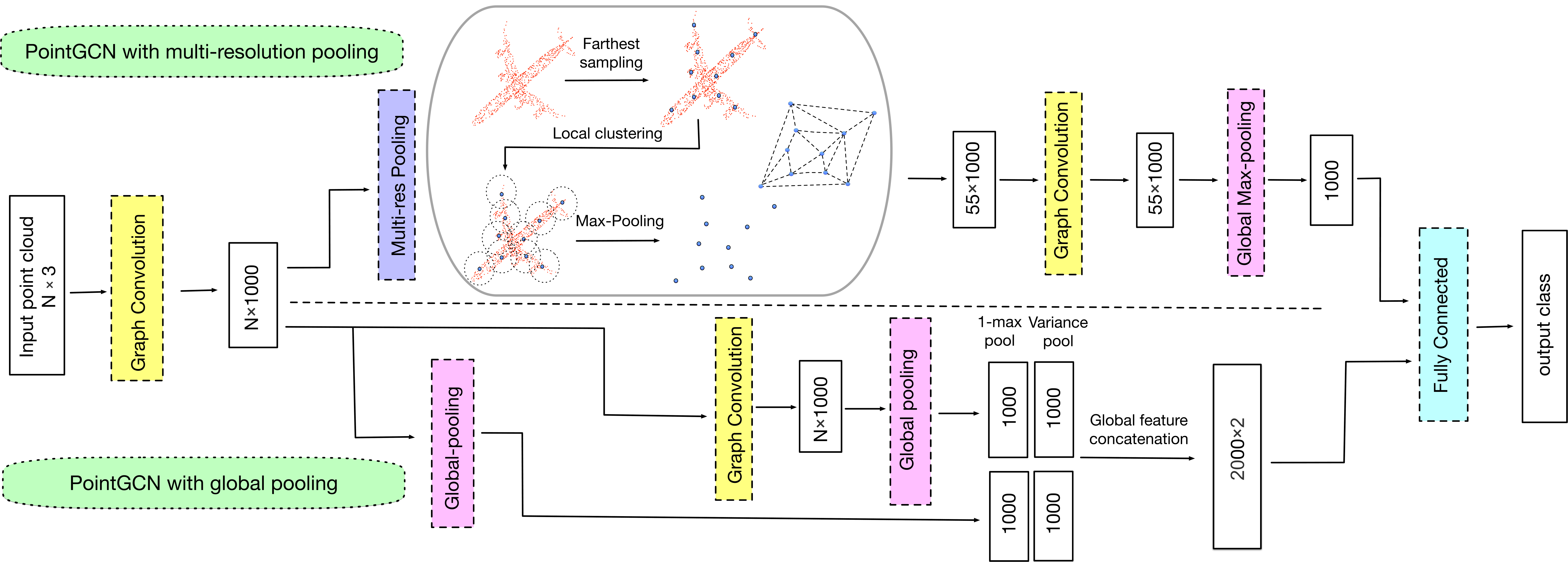}
\caption{Overall architecture for graph-CNN based point cloud classification model PointGCN. The top branch is the model architecture using multi-resolution pooling and the bottom branch is the model architecture using merely the global pooling layer.}
\label{overall_architecture}
\end{figure*}

Next we discuss our Graph-CNN architecture for 3D point cloud classification. Similar to typical CNNs and other Graph-CNN architectures, our architecture combines three main types of layers: convolutional, pooling, and fully-connected. Because the convolutional and pooling layers are particular to the graph setting, we describe these in detail. The overall architecture is shown in Figure \ref{overall_architecture}.

\textbf{Convolutional layer.}
During the training process, our goal is to train a set of graph filter coefficients that can translate the input signal to latent feature maps that capture relevant structure information to discriminate between object classes. Because we will apply the architecture to different graphs, and the Laplacian spectra of different graphs have different ranges, we first perform a normalization: we use the rescaled Laplacian $\tilde{L}=2L/\lambda_{\max}-I_n$, where $\lambda_{\max}$ is the largest Laplacian eigenvalue, so that all eigenvalues of $\tilde{L}$ are in the interval $[-1, 1]$. We have found that this improves stability as well as the performance of the network during learning.

As discussed in Sec.~3.1, each filter of order $K$ has $K$ learnable parameters (the filter coefficients), each parameter control the learned latent representation from different receptive field (from 1-hop neighbors to $K$-hops neighbors). Previous researchers have investigated the usage of the Chebyshev graph filtering approximation in distributed signal processing \cite{shuman_DCOSS_2011}, Graph-CNNs (e.g., ChebyNet~\cite{DBLP:conf/nips/DefferrardBV16}). Defferrard et al.~\cite{DBLP:conf/nips/DefferrardBV16} demonstrate the effectiveness of this convolution block in homogeneous graph prediction tasks such as image classification, where each images can be regarded as a grid graph (hence, having the same graph structure, assuming the image size is fixed in advance). 
We adapt a similar scheme to deal with heterogeneous graphs. Specifically, to obtain one level of feature transformation, we apply Chebyshev polynomial filters \eqref{cheby} and use the coordinate vectors $\bm{x}^{(1)}, \bm{x}^{(2)}, \bm{x}^{(3)}$ as input vectors for first convolution layer.

After each convolutional layer we apply a rectified linear unit (ReLu) nonlinear activation function. One of the advantages of the Chebyshev polynomial filtering is the learned feature maps will be localized within $K$-hops neighbors of each point. Besides, every time we add one more graph convolutional layer, the receptive field is enlarge by $K$ hops. 

\textbf{Pooling layer.}
The feature maps output by the convolutional layer are a point-wise latent representation. We implement two forms of pooling operations to aggregate information from these representations. One form of pooling computes global statistics across all output points, while the other form of pooling acts locally within the point cloud, leading to multi-resolution pooling similar to graph coarsening methods employed in previous work~\cite{DBLP:journals/corr/BrunaZSL13,DBLP:conf/nips/DefferrardBV16}.

\textit{Global pooling.} As discussed in Sec.~3.2, the output of graph filters can emphasize meaningful structural information about a point cloud. We seek representations that are invariant to point order permutation and rotations (to deal with the case when different point clouds have not been oriented/registered). To this end we use 1-max-pooling (i.e, taking the max of all filter outputs) and variance pooling (i.e, calculating the variance of each filter output across the $n$ points). Max pooling highlights the most distinctive points, whereas variance pooling quantifies spread of the outputs after filtering. 
When using multiple convolutional layers, we compute these statistics for each layer and concatenate them as inputs to the final layer for computing the final class probabilities. 

\textit{Multi-resolution pooling.} Graph clustering algorithms are computationally demanding, and we seek a light-weight method for local pooling of graph signals. In the special case of point cloud data, we can leverage the geometry inherent to the data to avoid explicitly calculating a graph coarsening or clustering. Instead, we realize multi-resolution pooling by sub-sampling a set of points that are most scattered from each other. This is done by first sampling a random point and adding it to the sampling set. The next sampling point is taken to be the one furthest from the initial sampling point, and subsequent sampling points are those that are mutually furthest from all other sampling points. After reaching the desired number of sampling points (resolution at the next level) we associate each non-sampled point with the nearest sampled point and aggregate among these groups by max-pooling.

\textbf{Final architecture.} The final architecture we use in the experiments below is composed of two graph convolutional layers. We construct a 40-nearest neighbor graph for each point cloud object to perform graph convolution. Each layer involves graph convolution with order $K=3$, followed by ReLu activation and one form pooling (we report the results of different combinations below). Both first and second convolutional layer has 1000 filters. When using multi-resolution pooling we downsample to 55 points for the second layer where clusters are formed by 50 nearest neighbors of every centroid point. The second convolutional layer is followed by one final global pooling step. The intermediate representations produced by the convolutional layers are flattened into a vector and passed through a final fully-connected linear layer with softmax activation, where the number of outputs is the number of classes. Thus, the final output corresponds to a vector with entries giving the predicted probability of the input belonging to each of the possible classes.

\textbf{Training details.}
 The training is done using Adam optimizer with mini-batch training (batch size of 28). To prevent the model from overfitting, two methods have been used. One is adding dropout after both convolutional layers and fully connected layer with 0.9 and 0.5 dropout rate respectively, and the other one is adding a $\ell_2$ regularization term (so-called “weight decay”) with coefficient $2 \times 10^{-4}$ to avoid learning complicated model.
 

\section{Performance Evaluation}
\label{sec:performance}

\textbf{Dataset description and preparation.}
We evaluate our algorithm on the ModelNet40 and ModelNet10 datasets for 3D object recognition~\cite{DBLP:conf/cvpr/WuSKYZTX15}. ModelNet10 contains 4,899 CAD models from 10 categories, split into 3,991 for training and 908 for testing. ModelNet40 contains 12,311 CAD models from 40 categories, split into 9,843 for training and 2,468 for testing. We use the same data format as PointNet~\cite{DBLP:journals/corr/QiSMG16}, where the data are uniformly sampled on the CAD object mesh face to obtain $n=2048$ points. Both datasets have unbalanced class distribution, which poses a challenge in model training. In ModelNet10 all models are oriented, and in ModelNet40 they are not oriented.

All the point clouds are initially normalized into a unit sphere. To further reduce the size of each object for fast computation, we preprocess the data to 1024 points per object by farthest subsampling.  We experimented with other preprocessing schemes, not reported here due to space limitations, such as contour-enhanced subsampling~\cite{DBLP:conf/icassp/ChenTFVK17}, but they didn't lead to any improvement in performance. However, using the contour-enhanced subsampling do lead to better resistance against data corruption.

\textbf{Performance comparison.} 
3D data have mainly three types of representations. Different representations of 3D objects lead to different approaches to solve the problem. We compare our algorithm with state-of-the-art methods using volume as input (i.e., binning into voxels) \cite{DBLP:conf/cvpr/WuSKYZTX15,DBLP:conf/iros/MaturanaS15,DBLP:journals/corr/GrahamM17}, using images from different views as input \cite{DBLP:conf/iccv/SuMKL15}, and using point sets \cite{DBLP:journals/corr/QiSMG16, DBLP:conf/cvpr/SimonovskyK17, DBLP:conf/nips/deepsets} as input respectively. Since the dataset has class imbalance, we consider two performance metrics, mean instance accuracy and mean category accuracy, to evaluate the performance. To address the class imbalance issue, we use the standard approach of weighting the loss associated with each training instance inversely proportional to the frequency that the corresponding class of the training instance appears in the training set, penalizing mistakes on less common classes more heavily than on more common classes. 

The results are shown in Table 1, where we can see that there is still a gap between our method and the state-of-the-arts 3D object classification approach MVCNN~\cite{DBLP:conf/iccv/SuMKL15}, which uses ImageNet1K to pre-train and uses an extensive data augmentation scheme (observe 3D objects from 80 views) to cope with the data where the orientation is not aligned. In contrast, we use no pre-training and only use input data from one view. Comparing to other point-based methods, on ModelNet 10 and ModelNet 40, we improve the performance compared to PointNet~\cite{DBLP:journals/corr/QiSMG16} and ECC~\cite{DBLP:conf/cvpr/SimonovskyK17} in both performance metrics.

\begin{figure}[t]
\centering
\includegraphics[width=6.5cm]{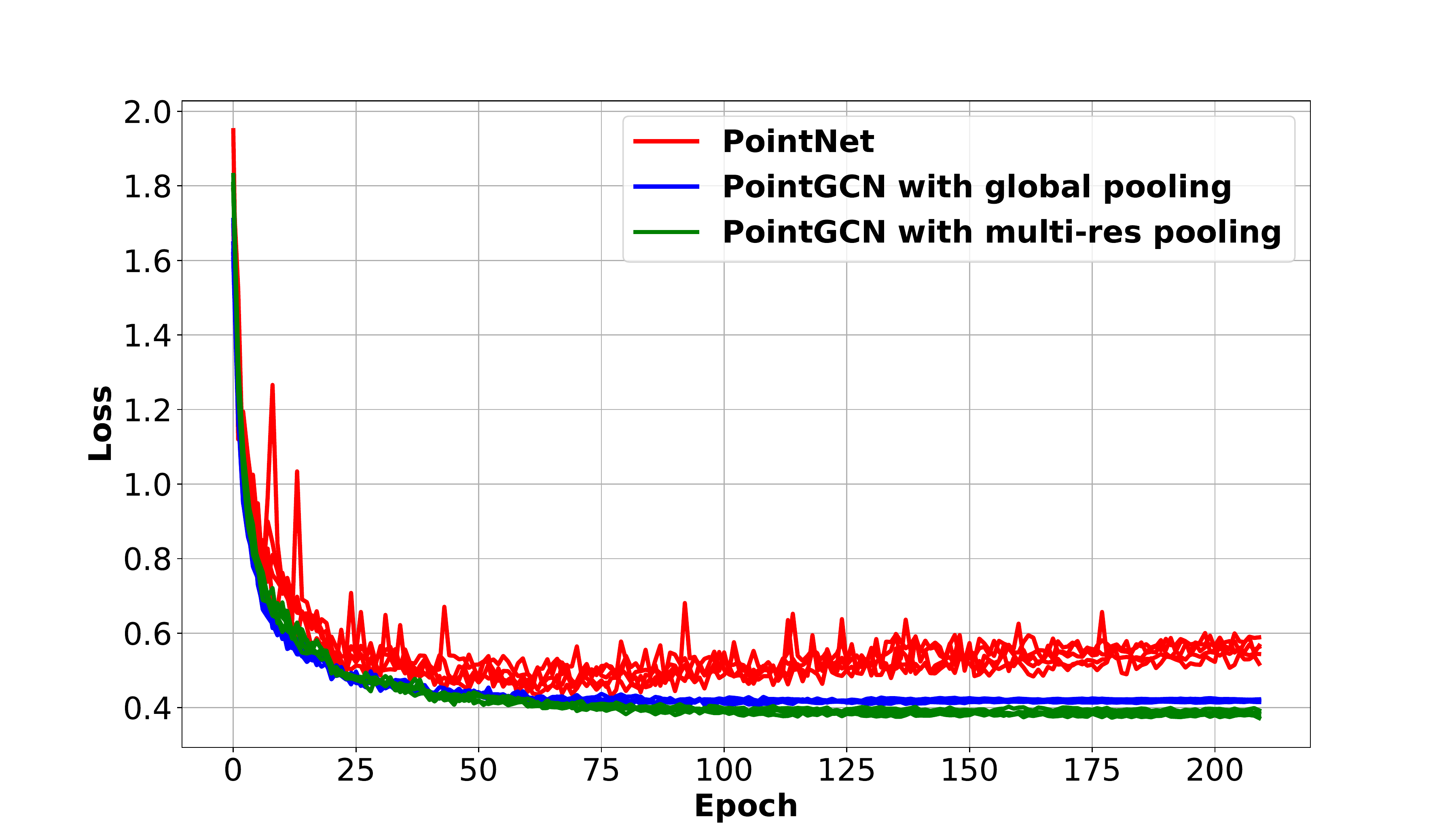}
\caption{Test set loss comparison for PointGCN and PointNet from five complete training process}
\label{fig:convergence_curve}
\end{figure}

\begin{figure}[t]
\centering
\includegraphics[width=6cm]{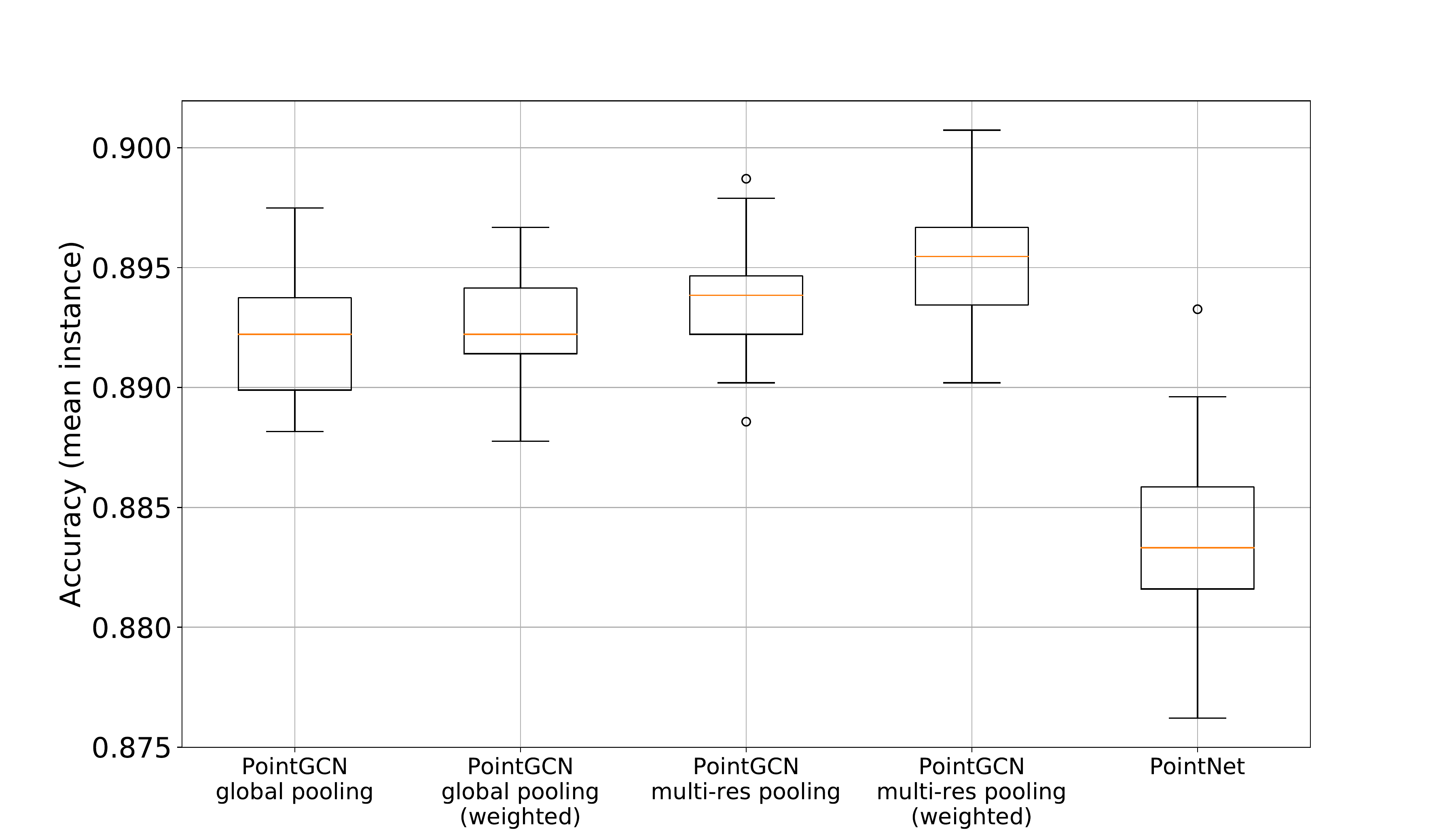}
\caption{Model stability comparison with PointNet and PointGCN from 50 trials (present results of our proposed model with and without weighting scheme for fair comparison).}
\label{fig:std_comparison}
\end{figure}

Fluctuation of the performance during the training (i.e., sensitivity to initial weights) is one of the difficulties when training on ModelNet40. This phenomenon is especially severe for PointNet~\cite{DBLP:journals/corr/QiSMG16}, demonstrated by the learning curve in Figure~\ref{fig:convergence_curve} and the model stability in Figure~\ref{fig:std_comparison}. Our proposed method has relatively faster convergence rate as well as improves the reliability of the model performance for both pooling approaches. One possible reason could be that PointNet only seek the latent representation in a global fashion and lack of exploration of local point features. And in our model, since we introduce the structure of the data by providing the local interconnection between points and explore graph features from different abstraction levels by the localized graph convolutional layers, it guarantees the exploration of more distinctive latent representations for each object class.

\begin{table}[]
\centering
{\tiny
\begin{tabular}{@{}cccccc@{}}
\toprule
Algorithm                                                                     & \begin{tabular}[c]{@{}c@{}}Input \\ format\end{tabular}         & \begin{tabular}[c]{@{}c@{}}ModelNet 10\\ Accuracy\\ (avg. class)\end{tabular} & \begin{tabular}[c]{@{}c@{}}ModelNet 10\\ Accuracy\\ (overall)\end{tabular} & \begin{tabular}[c]{@{}c@{}}ModelNet 40\\ Accuracy\\ (avg. class)\end{tabular} & \begin{tabular}[c]{@{}c@{}}ModelNet 40\\ Accuracy\\ (overall)\end{tabular} \\ \midrule
3D ShapeNets                                                                  & \begin{tabular}[c]{@{}c@{}}volume\\ (1 view)\end{tabular}       & 83.5\%                                                                        & -                                                                          & 77\%                                                                          & 84.7\%                                                                     \\
VoxNet                                                                        & \begin{tabular}[c]{@{}c@{}}volume\\ (12 views)\end{tabular}     & 92.0\%                                                                        & -                                                                          & 83\%                                                                          & 85.9\%                                                                     \\
SSCN                                                                        & \begin{tabular}[c]{@{}c@{}}volume\\ (20 views)\end{tabular}     & -                                                                             & -                                                                          & 88.2\%                                                                        & -                                                                          \\
MVCNN                                                                         & \begin{tabular}[c]{@{}c@{}}image\\ (80 views)\end{tabular}      & -                                                                             & -                                                                          & \textbf{90.1\%}                                                                        & -                                                                          \\ \midrule
ECC                                                                           & \begin{tabular}[c]{@{}c@{}}1024 points\\ (12 views)\end{tabular} & 90.0\%                                                                        & 90.8\%                                                                     & 83.2\%                                                                        & 87.4 $\pm$ 0.40\%                                                                     \\

PointNet                                                                    & \begin{tabular}[c]{@{}c@{}}1024 points\\ (1 view)\end{tabular}  & 91.53\%                                                                       & 91.74\%                                                                    & 85.35\%                                                                       & 88.37$\pm$0.33\%                                                                    \\
\begin{tabular}[c]{@{}c@{}}\textcolor{blue}{PointGCN}\\ \textcolor{blue}{(global pooling)}\end{tabular}           & \begin{tabular}[c]{@{}c@{}}\textcolor{blue}{1024 points}\\ \textcolor{blue}{(1 view)}\end{tabular}  & \textcolor{blue}{91.39\%}                                                                       & \textcolor{blue}{91.77\%}                                                                    & \textcolor{blue}{86.19\%}                                                                       & \textcolor{blue}{89.27$\pm$0.20\%}                                                                    \\
\begin{tabular}[c]{@{}c@{}}\textcolor{blue}{PointGCN}\\ \textcolor{blue}{(multi-resolution pooling)}\end{tabular} & \begin{tabular}[c]{@{}c@{}}\textcolor{blue}{1024 points}\\ \textcolor{blue}{(1 view)}\end{tabular}  & \textcolor{blue}{91.57\%}                                                                         & \textcolor{blue}{91.91\%}                                                                    & \textcolor{blue}{86.05\%}                                                                       & \textcolor{blue}{89.51$\pm$0.23\%}                                                                    \\

Deep Sets                                                                     & \begin{tabular}[c]{@{}c@{}}1000 points\\ (1 view)\end{tabular}  & -                                                                       & -                                                                   & -                                                                     &{87 $\pm$ 1\%}                                                                    \\
Deep Sets                                                                     & \begin{tabular}[c]{@{}c@{}}5000 points\\ (1 view)\end{tabular}  & -                                                                       & -                                                                   & -                                                                     & \textbf{90 $\pm$ 0.3\%}                                                                    \\

\bottomrule
\end{tabular}}
\label{acc_comparison}
\caption{Results comparison with state-of-art methods on ModelNet.  *Means the result is reproduced by the code provided by the paper author. Otherwise, the baseline results presented here are reported in the original paper. All the results reported above for our proposed method and reproduced method are the average results of 50 trials for each scenario.}
\end{table}

\begin{figure}[t]

\begin{minipage}[b]{0.31\linewidth}
  \centering
  \centerline{\includegraphics[width=1.9cm]{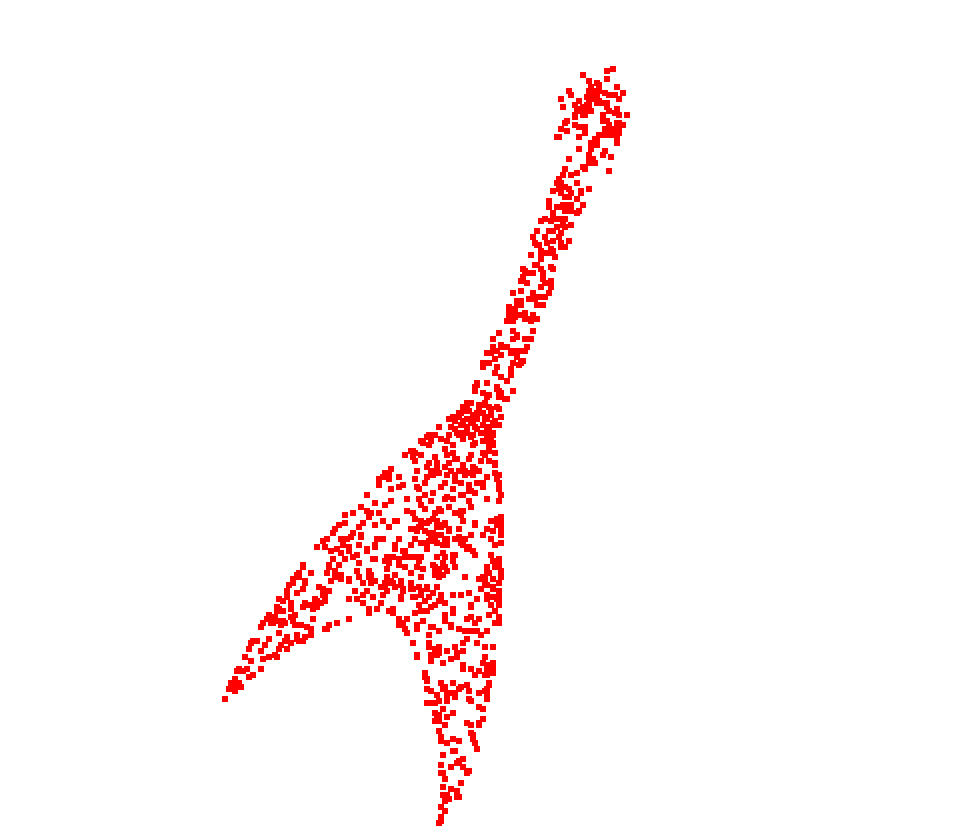}}
\end{minipage}
\begin{minipage}[b]{0.31\linewidth}
  \centering
  \centerline{\includegraphics[width=1.9cm]{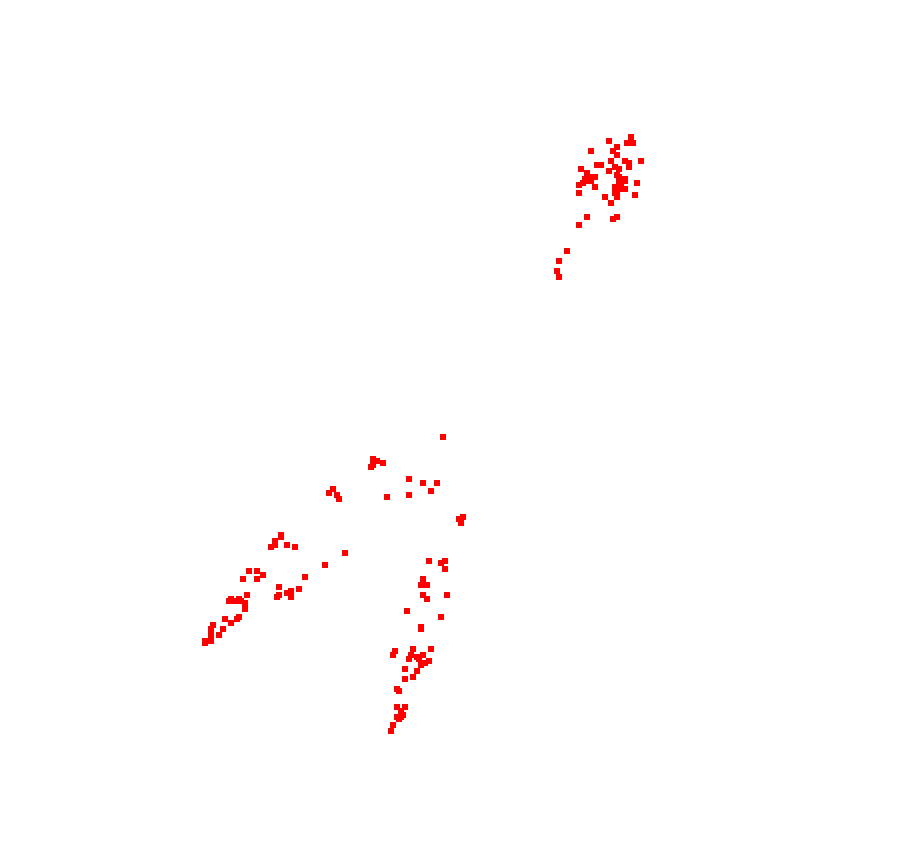}}
\end{minipage}
\vspace{0cm}
\begin{minipage}[b]{0.31\linewidth}
  \centering
  \centerline{\includegraphics[width=1.9cm]{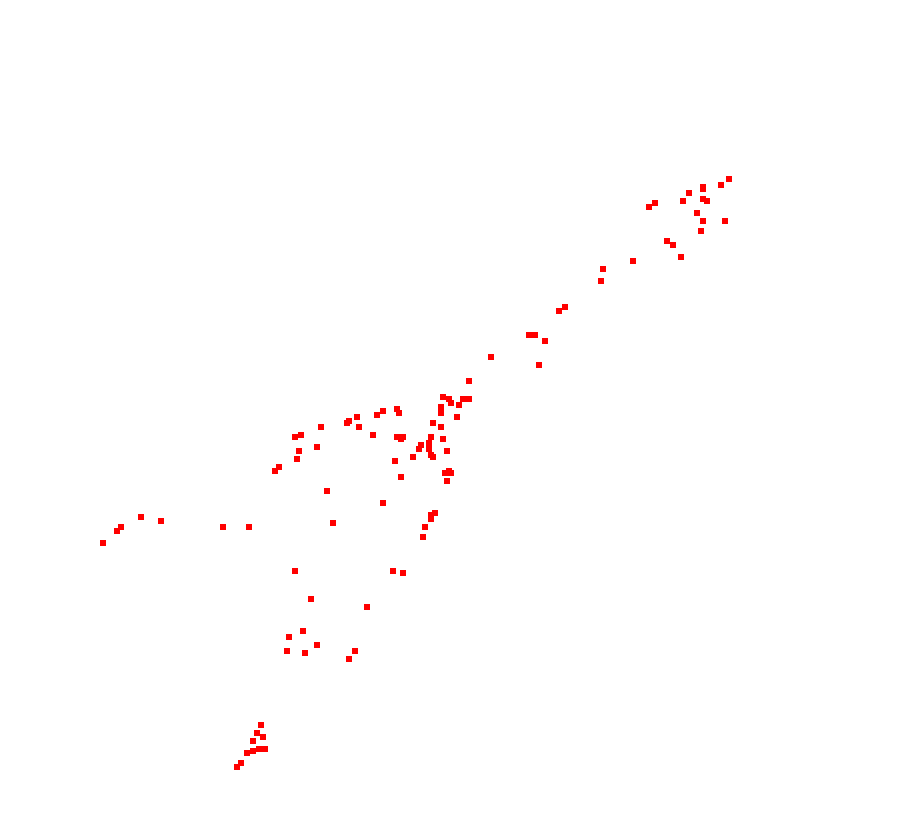}}
  
\end{minipage}

\begin{minipage}[b]{0.31\linewidth}
  \centering
  \centerline{\includegraphics[width=1.9cm]{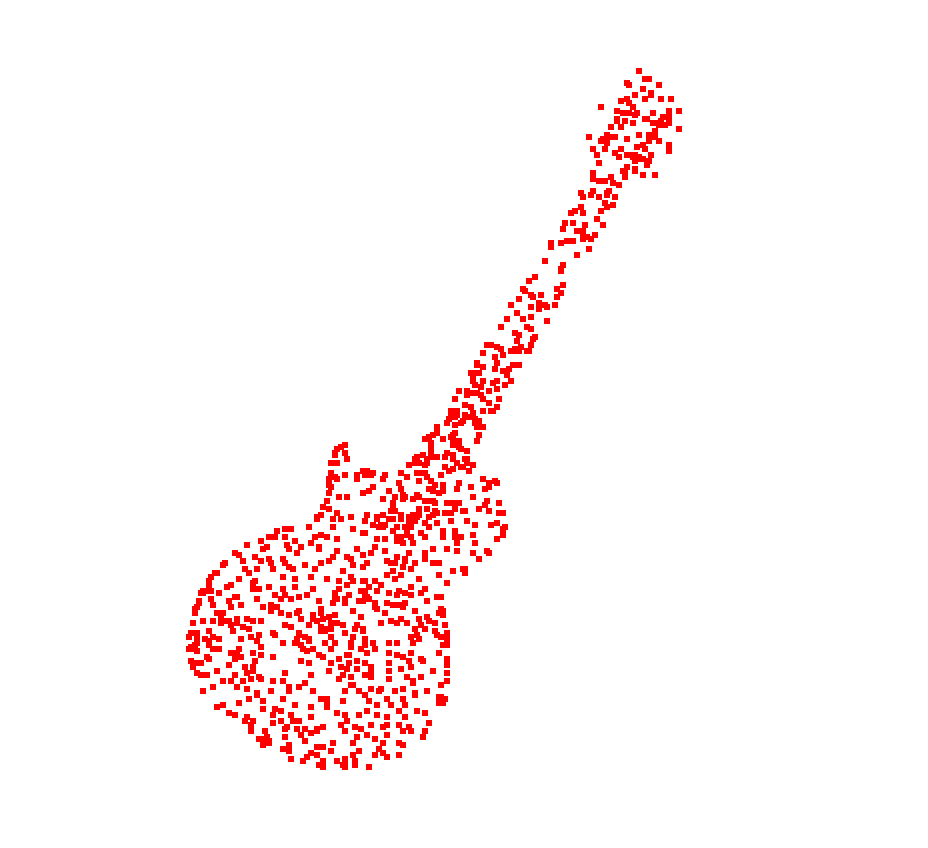}}
\end{minipage}
\begin{minipage}[b]{0.31\linewidth}
  \centering
  \centerline{\includegraphics[width=1.9cm]{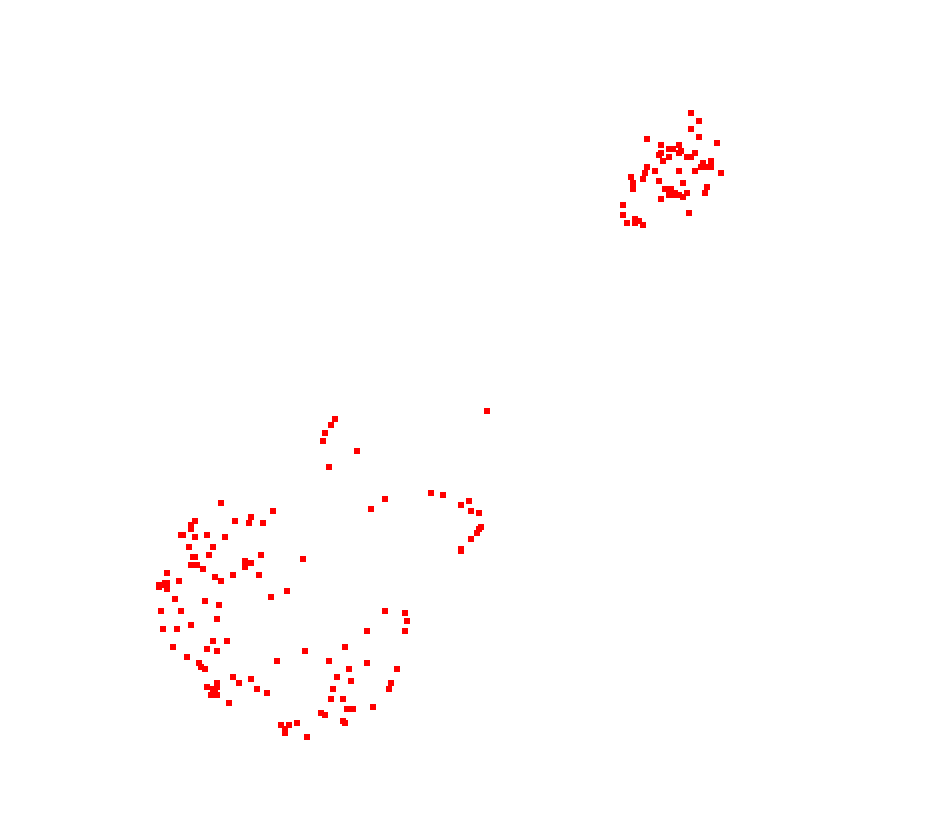}}
\end{minipage}
\vspace{0cm}
\begin{minipage}[b]{0.31\linewidth}
  \centering
  \centerline{\includegraphics[width=1.9cm]{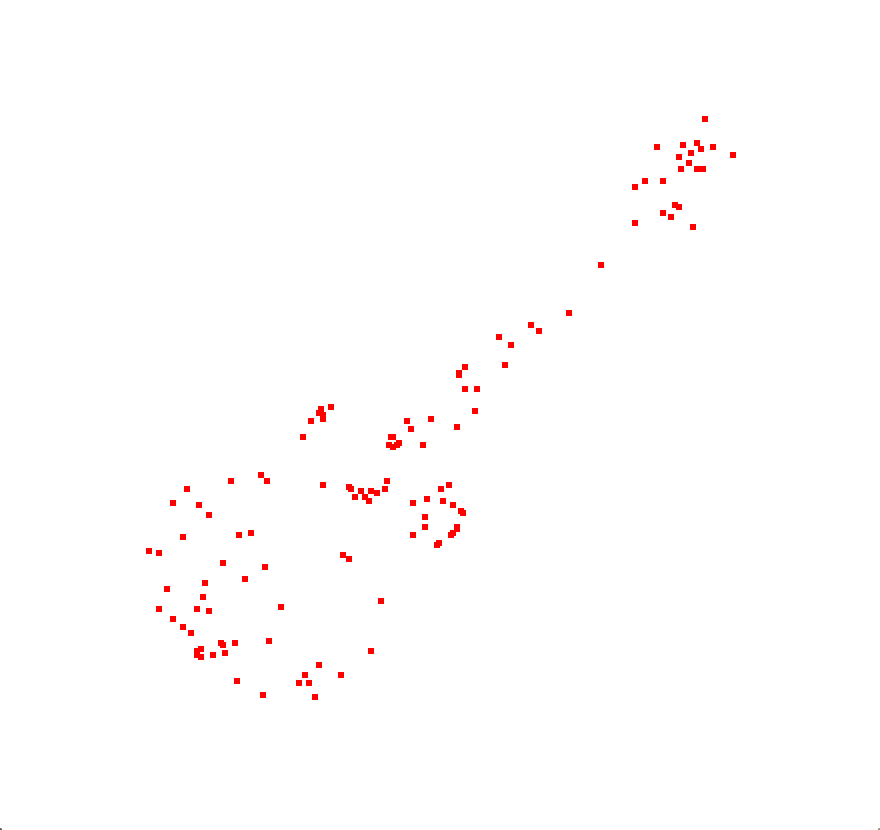}}
  
\end{minipage}

\begin{minipage}[b]{0.31\linewidth}
  \centering
  \centerline{\includegraphics[width=1.9cm]{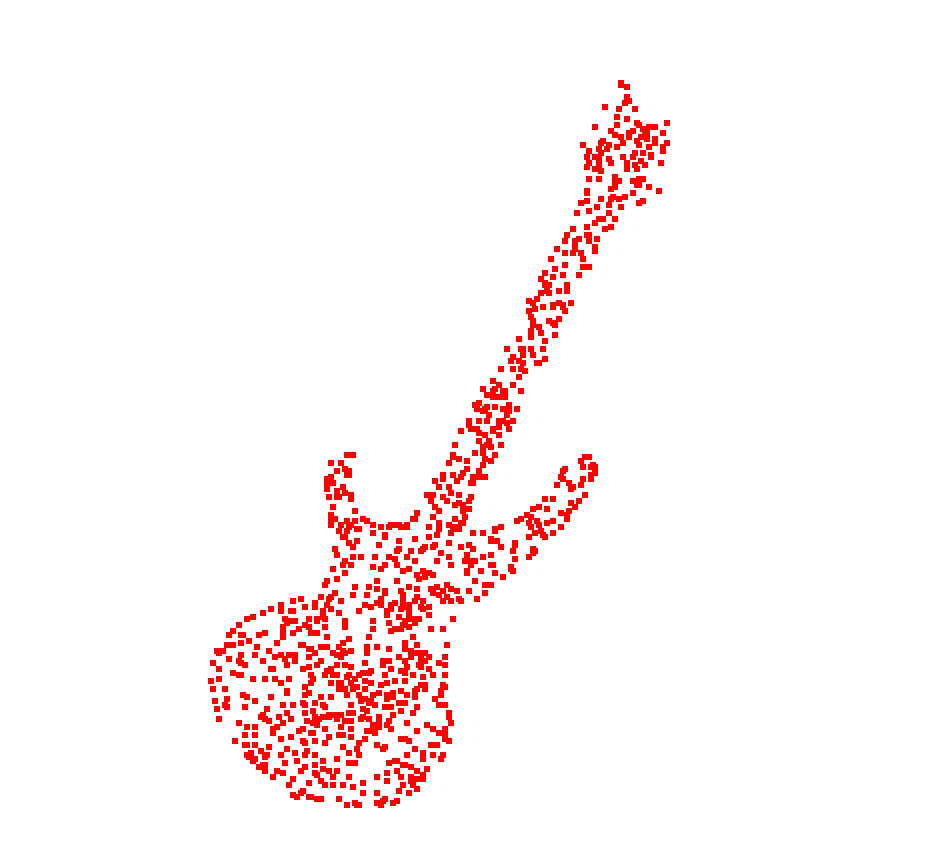}}
\end{minipage}
\begin{minipage}[b]{0.31\linewidth}
  \centering
  \centerline{\includegraphics[width=1.9cm]{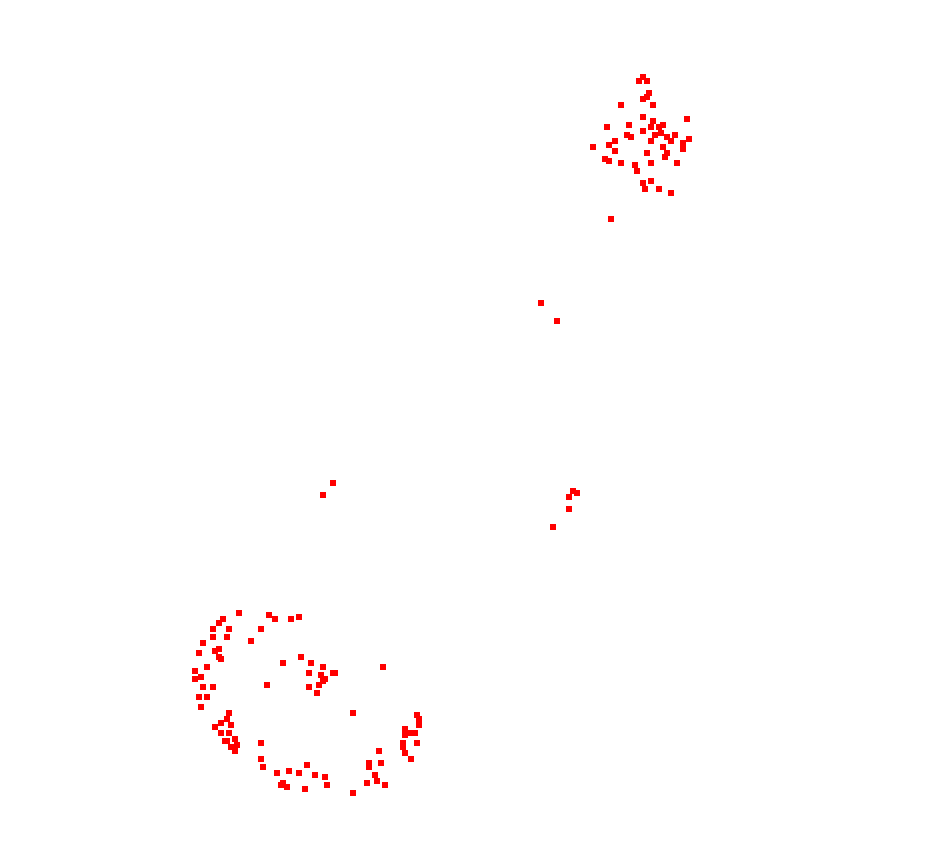}}
\end{minipage}
\vspace{0cm}
\begin{minipage}[b]{0.31\linewidth}
  \centering
  \centerline{\includegraphics[width=1.9cm]{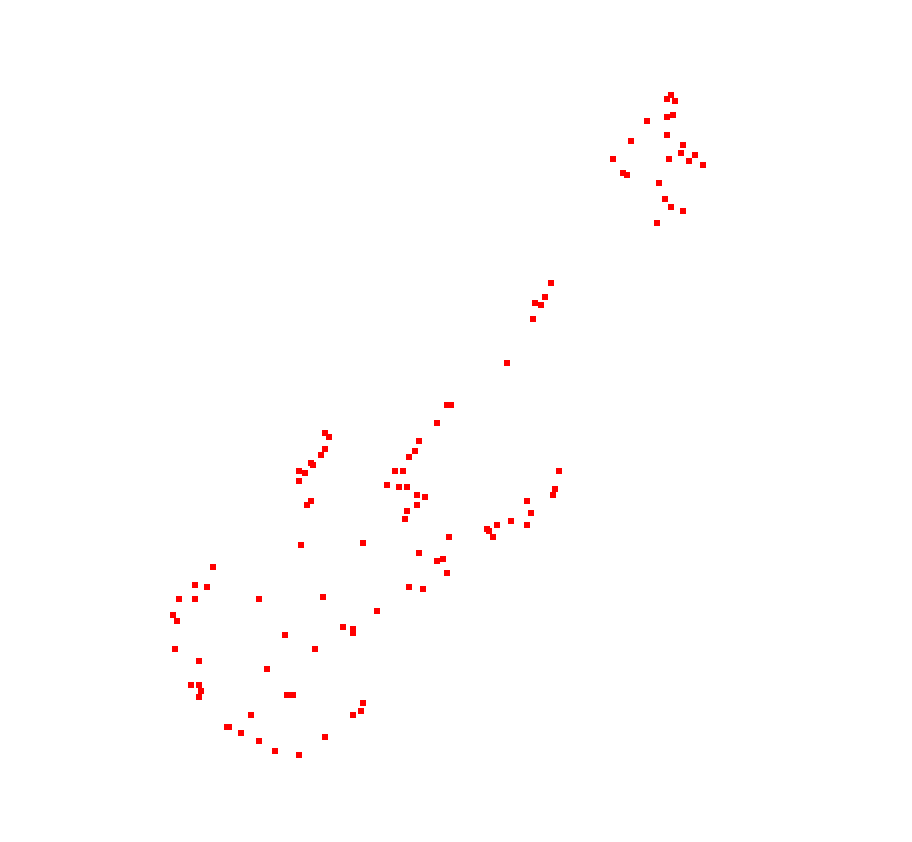}}
\end{minipage}

\caption{\label{fig:max_pooling_guitars} Max pooling contribution points from different layers (Guitar); \textit{Left:} Original point cloud; \textit{Middle:} Layer one active points; \textit{Right:} Layer two active points}
\label{fig:res}
\end{figure}

\textbf{Visualizing the effect of max-pooling.}
One of the key operations in the proposed architecture is the global operation max pooling, which aims to pick the unique pattern points and summarize the global signature of each object. We call the points that have the maximum values among each feature map the active points. We visualize an example of these active points for feature maps coming from both graph convolutional layers in Fig.~\ref{fig:max_pooling_guitars}. 
Active points at the first layer appear to emphasize local patterns, while those at the second layer emphasize more global structural information, which is consistent with our assumption that the $K$-localized graph convolutional layer can explore features from different receptive fields. 

\section{Conclusion}
\label{sec:conclusion}
In this paper, we propose a Graph-CNN model for 3D point cloud classification. The model has two fast localized graph convolutional layers and point cloud data specific designed pooling layer using global pooling or multi-resolution pooling. Our proposed approach is demonstrated to be competitive on the 3D point-cloud classification benchmark dataset ModelNet~\cite{DBLP:conf/cvpr/WuSKYZTX15}.

The proposed method has a number of interesting properties. First, by leveraging geometric information encoded in the graph structure, we narrow the search space for the learned model, which makes the model converge faster and also improves the robustness (as illustrated via the standard deviation of the model performance). Second, one of the biggest problems for point-based classification problem methods is how to achieve point order invariant. 
In the proposed algorithm, the features we learn are spatially localized by design since filters of order $K$ combine information from $K$-hop neighbors and order permutation preserves nearest neighbors, which are encoded in the constructed graph.
At last, because the proposed approach operates on graphs which are symmetric by design, the resulting filters (defined in terms of the Laplacian) are isotropic and do not capture any notion of directionality along the manifold from which the points were sampled. This property guarantees that the learned model is robust under rotation transformation, which explains that our proposed model perform well in ModelNet 40 (objects orientation are not aligned) without providing the input point cloud data from different views.

\bibliographystyle{IEEEbib}
\bibliography{Main} 

\end{document}